\documentclass{article}

\usepackage{arxiv}

\usepackage[utf8]{inputenc} 
\usepackage[T1]{fontenc}    
\usepackage{hyperref}       
\usepackage{url}            
\usepackage{booktabs}       
\usepackage{amsfonts}       
\usepackage{nicefrac}       
\usepackage{microtype}      
\usepackage{lipsum}
\usepackage{graphicx}
\graphicspath{ {./images/} }
\usepackage{listings}
\usepackage{color} 
\usepackage{tabularx}
\definecolor{codegreen}{rgb}{0,0.6,0}
\definecolor{codegray}{rgb}{0.5,0.5,0.5}
\definecolor{codepurple}{rgb}{0.58,0,0.82}
\definecolor{backcolour}{rgb}{0.95,0.95,0.92}

\lstdefinestyle{mystyle}{
	backgroundcolor=\color{backcolour},   
	commentstyle=\color{codegreen},
	keywordstyle=\color{magenta},
	numberstyle=\tiny\color{codegray},
	stringstyle=\color{codepurple},
	basicstyle=\ttfamily\footnotesize,
	breakatwhitespace=false,         
	breaklines=true,                 
	captionpos=b,                    
	keepspaces=true,                 
	numbers=left,                    
	numbersep=5pt,                  
	showspaces=false,                
	showstringspaces=false,
	showtabs=false,                  
	tabsize=2
}

\lstdefinelanguage{json}{
	basicstyle=\normalfont\ttfamily,
	numbers=left,
	numberstyle=\scriptsize,
	stepnumber=1,
	numbersep=8pt,
	showstringspaces=false,
	breaklines=true,
	frame=lines,
	backgroundcolor=\color{backcolour},
	stringstyle=\color{codepurple},
	literate=
	*{0}{{{\color{codepurple}0}}}{1}
	{1}{{{\color{codepurple}1}}}{1}
	{2}{{{\color{codepurple}2}}}{1}
	{3}{{{\color{codepurple}3}}}{1}
	{4}{{{\color{codepurple}4}}}{1}
	{5}{{{\color{codepurple}5}}}{1}
	{6}{{{\color{codepurple}6}}}{1}
	{7}{{{\color{codepurple}7}}}{1}
	{8}{{{\color{codepurple}8}}}{1}
	{9}{{{\color{codepurple}9}}}{1}
	{:}{{{\color{codepurple}:}}}{1}
	{,}{{{\color{codepurple},}}}{1}
	{\{}{{{\color{codepurple}\{}}}{1}
	{\}}{{{\color{codepurple}\}}}}{1}
	{[}{{{\color{codepurple}[}}}{1}
	{]}{{{\color{codepurple}]}}}{1},
}

\title{AI Multi-Agent Interoperability\\[1ex] \large Extension for Managing Multiparty Conversations}

\author{ \href{https://orcid.org/0009-0008-7513-1255}{\includegraphics[scale=0.06]{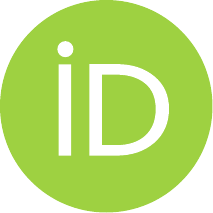}\hspace{1mm}Diego Gosmar} \\
	Chief AI Officer XCALLY\\
	Open Voice Interoperability Initiative Member\\
	Linux Foundation AI \& Data\\
	Torino, TO 10100, Italy \\
	\texttt{diego.gosmar@ieee.org} \\
	\And
	\href{https://orcid.org/0000-0002-3389-2784}{\includegraphics[scale=0.06]{orcid.pdf}\hspace{1mm}Deborah A. Dahl} \\
	Principal Conversational Technologies\\
	Open Voice Interoperability Initiative Member\\
	Linux Foundation AI \& Data\\
	Plymouth Meeting, Pennsylvania, USA \\
	\texttt{dahl@conversational-technologies.com} \\
	\And
	\href{https://orcid.org/0009-0001-3770-4963}{\includegraphics[scale=0.06]{orcid.pdf}\hspace{1mm}Emmett Coin} \\
	Founder ejTalk\\
	Open Voice Interoperability Initiative Member\\
	Linux Foundation AI \& Data\\
	Bellville, Michigan, USA \\
	\texttt{emmett@ejtalk.com} \\
	\And
	\href{https://orcid.org/0009-0005-5161-8120}{\includegraphics[scale=0.06]{orcid.pdf}\hspace{1mm}David Attwater} \\
	Senior Research Scientist Talkmap\\
	Open Voice Interoperability Initiative Member\\
	Linux Foundation AI \& Data\\
	Southport, Merseyside, United Kingdom \\
	\texttt{david.attwater@talkmap.com} \\
}

\begin{document}
\maketitle
\begin{abstract}
	This paper presents a novel extension to the existing Multi-Agent Interoperability specifications of the Open Voice Interoperability Initiative (originally also known as OVON from the Open Voice Network). This extension enables AI agents developed with different technologies to communicate using a universal, natural language-based API or NLP-based standard APIs. Focusing on the management of multiparty AI conversations, this work introduces new concepts such as the Convener Agent, Floor-Shared Conversational Space, Floor Manager, Multi-Conversant Support, and mechanisms for handling Interruptions and Uninvited Agents. Additionally, it explores the Convener's role as a message relay and controller of participant interactions, enhancing both scalability and security. These advancements are crucial for ensuring smooth, efficient, and secure interactions in scenarios where multiple AI agents need to collaborate, debate, or contribute to a discussion. The paper elaborates on these concepts and provides practical examples, illustrating their implementation within the conversation envelope structure.
\end{abstract}

\keywords{Conversational AI \and Artificial intelligence \and AI Interoperability \and Multi-Agency \and Agentic AI \and Agentive AI \and Chatbot \and Voicebot \and Intelligent Assistant \and Multiparty AI \and AI Conference Specifications}

\section{Introduction}

\subsection{Previous work}
There is increasing recognition that many applications of conversational systems can be addressed more successfully if the expertise required to perform the application is not expected to reside in one agent, but is allocated among independent agents with their own capabilities. In this approach, these independent agents contribute their individual knowledge to address specific aspects of the problem and communicate to provide their input toward achieving the overall goal.  It is clear that this perspective broadly resembles the traditional and very successful object-oriented programming paradigm and shares similar advantages, such as encapsulation, maintainability, and scalability. \\
The integration of multiple agents has been addressed in many approaches, which vary in the degree of autonomy of each agent, the degree of architectural similarity expected of the agents, and whether or not the agents have to be known ahead of time when the system is developed. We believe that the most useful systems will be those that maximally encapsulate the agents’ functionality, limit dependencies on specific semantic formats, and which can be configured dynamically at runtime. In addition, multi-agent systems that are based on proprietary frameworks are, by definition, constrained to coordinating agents that are based on those frameworks, automatically ruling out the possibility for the thousands of existing legacy conversational systems to participate. \\
We focus here on previous work aimed at coordinating full agents as opposed to work aimed at coordinating specific modality components, such as the W3C Multimodal Architecture \cite{w3c} and the Galaxy Communicator Software Infrastructure \cite{galaxy}.

\subsubsection{Early systems}
Knowledge Query Manipulation Language (KQML) \cite{KQML} was an early system that focused on knowledge sharing and communication among intelligent agents. While it supported cooperation among intelligent agents in multi-agent systems, it relied on shared ontological assumptions among agents, which created a barrier to deployment. In addition, KQML did not specifically address conversational interactions and was more focused on sharing knowledge among agents.
VoiceXML \cite{vxml} was another early approach to collaboration among agents. It allowed conversations to be passed to other Voice-XML based agents through the <transfer> element. However, VoiceXML required agents to be based on VoiceXML, conversations could only be transferred to a single receiving agent, and no previous conversational context could be passed to the receiving agent.
Systems like the Open Agent Architecture \cite{oaa} used Inter-Agent Communication Languages to facilitate collaboration among independent agents. While this reduced dependencies on specific internal architectures, it required agents to interpret highly structured semantic representations rather than natural language, which constrained the flexibility and scalability of the system.

\subsubsection{Recent systems}
The emergence of very capable LLMs has led to a dramatic increase in both research and deployed conversational systems. Along with this increase, the value of multi-agent systems is becoming more apparent,  and a number of new multi-agent frameworks have recently become available. Some examples include the following:
\begin{itemize}
	\item \textbf{OpenAI Swarm} \cite{swarm}: A framework designed to orchestrate multiple AI agents collaboratively to accomplish complex tasks. Released with Open Source MIT License, it is currently considered to be an experimental system that is not ready for deployment. In particular, Swarm is currently an experimental sample framework intended to explore ergonomic interfaces for multi-agent systems. It is not intended to be used in production, and therefore has no official support.
	\item \textbf{Microsoft Autogen}: An Open-Source Programming Framework for Agentic AI. AutoGen is powered by collaborative research studies from Microsoft, Penn State University, and University of Washington and licensed under the Creative Commons Attribution 4.0 International \cite{autogen}.
	\item \textbf{CrewAI}: A platform that orchestrates multiple AI models and services to perform cohesive workflows. It emphasizes flexibility in integrating diverse AI technologies and managing complex task sequences. \cite{crewai}.
	\item \textbf{Multi-Agent Orchestrator framework}: a tool by Amazon AWS Labs for implementing sophisticated AI systems comprising multiple specialized agents. Its primary purpose is to intelligently route user queries to the most appropriate agents while maintaining contextual awareness throughout interactions \cite{awslabs}. The project is Open Source and provides optimal integration and performance combined with the Amazon AWS Bedrock fully managed service \cite{bedrock}.
	\item \textbf{Mixture of Experts (MoE) Paradigm} \cite{mixture}: While not an architecture for multi-agent systems per se, the Mixture of Experts paradigm represents another example of an architecture where a larger system relies on encapsulated individual components that supply complementary expertise with respect to an overall problem. 
\end{itemize}
We will see in the following sections that, in fact, these multi-agent frameworks could be compatible with the framework presented in this paper. By providing any of them with an OVON wrapper, they could become interoperable with other OVON-compatible systems. For instance, an Amazon Bedrock service with multiple internal agents could externally present itself as a black box, communicating with other multi-agent frameworks (e.g., Microsoft Autogen, CrewAI, or other platforms not listed here) by leveraging the OVON Universal API specifications.

\subsection{The Initial OVON Framework}
In contrast to other approaches, the OVON (Open Voice Network) framework introduced in our previous work\cite{ovonspec} sought to overcome some interoperability limitations by establishing a highly scalable and flexible method for AI agent interoperability\footnote{For the remainder of this document, the term "agent" will be used to refer to an entity with the capacity to act, while "agency" or “agentic” will denote the exercise or manifestation of this capacity, in accordance with the definition provided by Markus Schlosser\cite{plato2015}.}. Our framework supports a wide range of independent assistants, regardless of their underlying technologies, enabling them to collaborate through minimal communication standards. This loose coupling dramatically reduces the complexity of integrating new assistants into the ecosystem, thereby enhancing scalability. 

\subsection{Enhancements to the OVON Framework}
However, this initial work covers only conversations between one user and one assistant at a time. That is, if the user wants to get information from more than one assistant, they have to access multiple assistants in sequence. This most likely will have two less-than-optimal consequences. In the first place, any information from the conversation with the first assistant that is required by the second assistant will have to be explicitly transferred to the second assistant when the second assistant is invited to the conversation. The second and more significant drawback is that any higher-level conclusions resulting from the various conversations will have to be determined by the user. That is, since the assistants don’t know about the other tasks, they won’t be able to make suggestions that combine information gathered from other assistants with their own information.

\subsection{Use Cases}
Let’s look at an example use case for managing multi-party conversations via a multi-agent AI. Suppose a user is planning a trip that involves booking a flight, a rental car, and a hotel, and also involves looking for interesting things to do in the destination city. This planning could involve conversations with four or more assistants at different travel services. The travel dates, which all of these assistants need, have to be passed to each assistant in turn to avoid making the user repeat them. In addition, if the assistants are talking together, the tourist information assistant could point out that there is a music festival that the user would enjoy, but attending it would require extending the trip by one day. If the tourist assistant is involved in the flight booking conversation, it could tell the user about the festival before the user books their flight. This could save the user a lot of time.

\begin{figure}[h!]
	\centering
	\includegraphics[width=\linewidth]{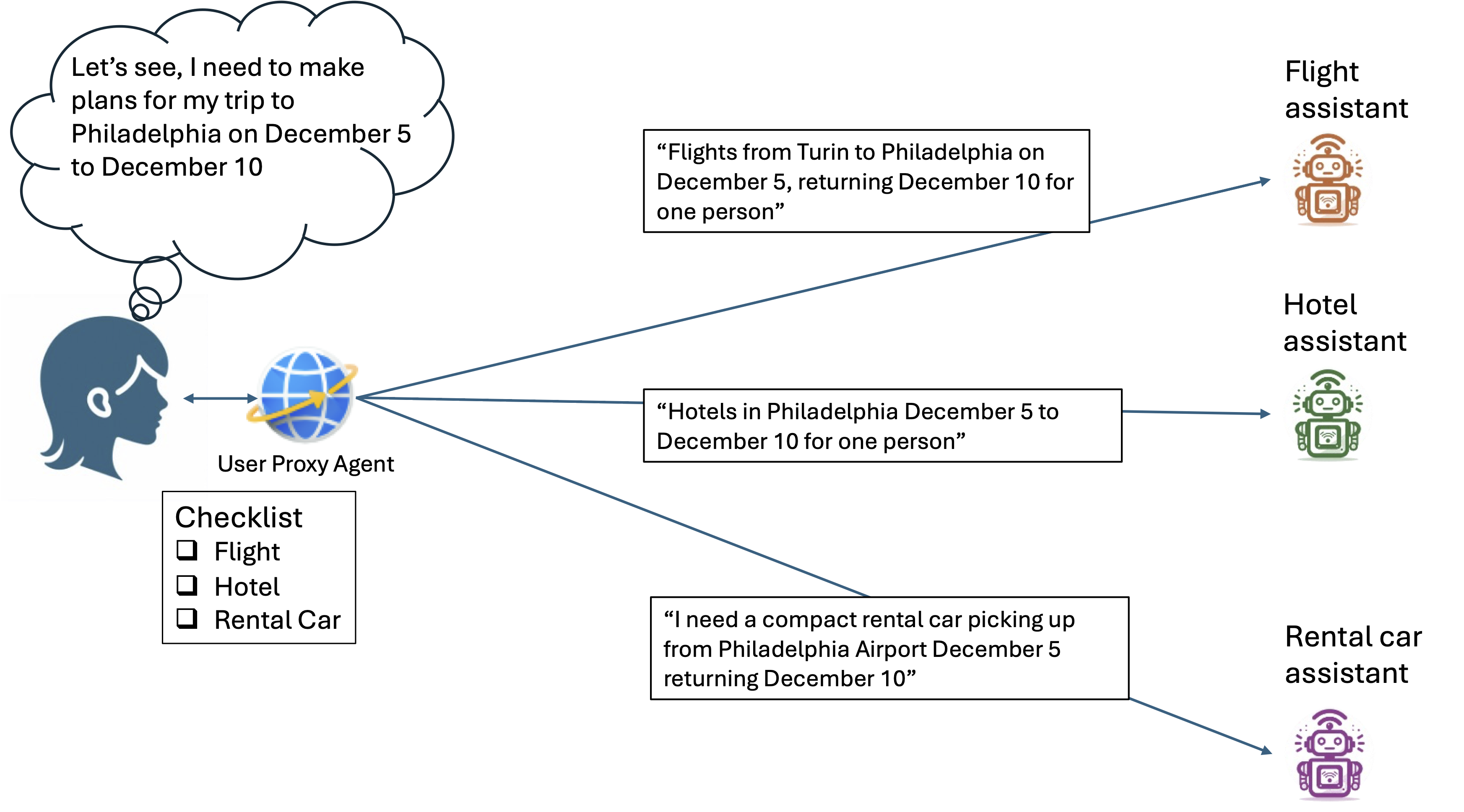}
	\caption{One assistant at a time}
	\label{fig:fig1}
\end{figure}

\begin{figure}[h!]
	\centering
	\includegraphics[width=\linewidth]{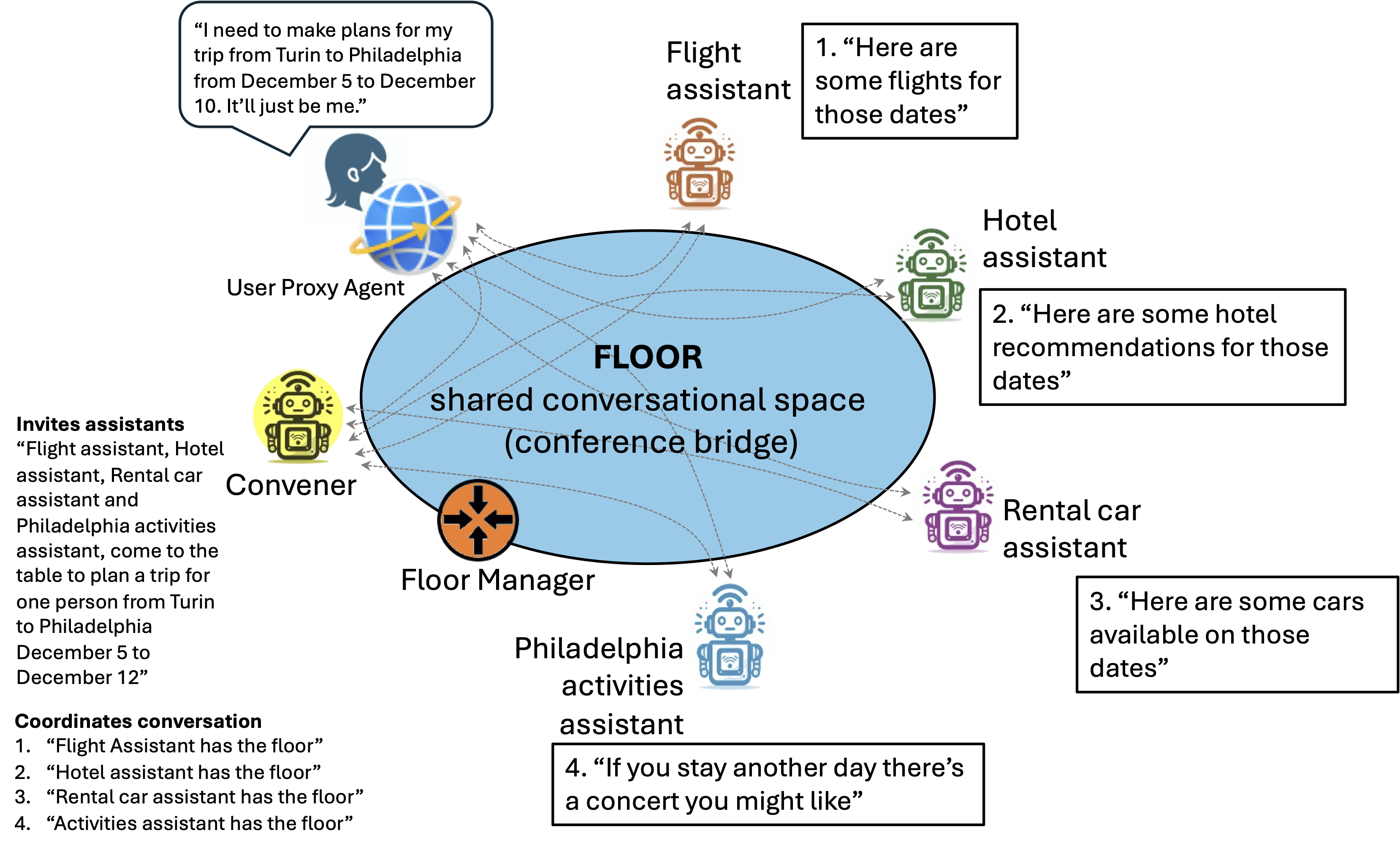}
	\caption{Multiple assistants in the same conversation}
	\label{fig:fig2}
\end{figure}

Figures \ref{fig:fig1} and \ref{fig:fig2} contrast these two situations. In Figure \ref{fig:fig1}, the user is planning a trip by accessing three assistants with different expertise in sequence. 

Here the user has to keep track of the tasks that she has to coordinate to plan her trip--flight, hotel, and rental car, and discusses her plans separately with each assistant, repeating some of the trip details in each conversation. This is the way that most planning with multiple assistants currently works.\\
In contrast, Figure  \ref{fig:fig2} shows multiple conversational agents gathered in a shared conversational space, referred to as the Floor, brought together by a Convener, with conversational turn-taking managed by the Convener itself. These concepts will be discussed more fully in the following sections of the paper.

A similar use case is described in \cite{multiframework}, where several agents are jointly assigned the task of allocating beds to hospital patients.The agents (nurse proxy, bed allocation specialist and patient database, among others) all have widely differing knowledge that makes it impractical to combine the agents into a single expert. Each agent has its own knowledge which it brings to the discussion of how to allocate a bed to a specific patient, arguing why or why not a particular bed is suitable for that patient. It would be very cumbersome if the user had to consult each agent in sequence to perform this task. Many other AI healthcare-specific applications could benefit from having conversational AI multi-agents coordinate with each other to enhance awareness of patient situations, including, for example, this risk detection model\cite{riskdet} for assisting vulnerable people.

\subsection{Requirements and proposed extensions}
For these reasons, we propose to extend the earlier two-party conversational specifications\cite{ovonspec} to handle requirements for conversations involving multiple assistants. Multi-party dialog systems have been discussed in the literature, for example \cite{multiframework}\cite{multirobot} among others. \cite{multirobot} describes a multi-agent system with user-initiative, where several agents can be present but the agents don’t collaborate -- they simply respond individually to user questions. \cite{multiframework} describes a system for collaborative problem-solving among agents, but it is restricted to one domain in that all of the agents are experts in different aspects of a larger problem. Our goal is to be able to support mixed-initiative applications with multiple agents that collaborate across domains.
These are the requirements that we propose for support of multi-party conversations:
\begin{enumerate}
	\item It must be possible to hold a conversation among more than two conversants.
	\item Conversants must be able to come and go during a conversation.
	\item It should be possible for a subset of conversants to be able to hold private conversations among themselves.
	\item There should be no fixed limit on the number of conversants.
	\item There should be a way to control possible unruly conversants through techniques like muting or ejecting.
\end{enumerate}
Requirement 1 is the key requirement for support of multi-party conversations. The other requirements support it.
This paper extends the initial specifications by introducing key concepts that address the specific requirements of managing multiparty conversations within the context of AI-driven multiparty conferences. The new concepts introduced in this work—such as the Floor, and related Multi-Conversant Support, Convener Agent, and mechanisms for handling Interruptions and Uninvited Agents—are designed to ensure that AI agents can collaborate effectively in dynamic, multi-agent environments. These extensions not only enhance the framework’s ability to handle complex, multi-party interactions but also ensure that the system can scale to accommodate a growing number of agents and tasks.
For instance, in scenarios where a human interacts with multiple AI assistants for various tasks—such as coordinating events, managing appointments, or retrieving information—the framework ensures effective communication and task delegation among the agents. This is achieved independently of each agent's underlying technologies or models, showcasing the system’s ability to scale across different applications and user needs.
Previous work\cite{convainteroperability} laid the foundation for AI agent interoperability, establishing the basic framework for seamless communication between independent conversational agents. However, the extensions presented in this paper are essential for overcoming the challenges associated with scalability and effective management in multiparty conversational settings. These enhancements introduce a versatile and adaptable platform that ensures AI-driven multiparty conferences can be conducted smoothly, with agents collaborating efficiently and effectively, regardless of their technological diversity. This approach not only addresses the current needs of evolving AI ecosystems but also provides a robust and future-proof solution capable of integrating new agents and capabilities as they emerge.

\section{Extensions to the Conversation Envelope}
The concept of Multi-Agent Interoperability revolves around creating a shared protocol, based on standard universal APIs using NLP, that allows heterogeneous AI agents to communicate effectively. This is achieved through a standardized conversation envelope API, as detailed in previous research\cite{ovonspec}, which defines the message structure and communication protocols. In multiparty scenarios, such as AI-driven conferences or collaborative tasks, the existing framework needs to be enhanced to manage the flow of conversation among multiple agents, handle interruptions, and secure the conversation from uninvited agents. These scenarios require additional layers of management that were not addressed in the initial framework.
To address these complex interactions, this paper introduces several key extensions to the conversation envelope framework, specifically designed to enhance the coordination and management of multiparty conversations. The next sections will dive into these extensions, including the introduction of a Floor Manager to transmit inter-agent messages, Multi-Conversant Support to enable collaboration among multiple agents via Natural language, the Convener Agent as a conversational coordinator, and mechanisms to handle interruptions and manage unwanted agents, ensuring that all interactions remain orderly and productive. Note that these extensions are backward-compatible with the basic conversation envelope messages\cite{ovonspec} and it is not necessary for systems to support them if the application doesn’t require multiple agents. 

\subsection{The Floor Shared Conversational Space}
In addition to the Convener, this paper introduces the concept of the "Floor", which can be thought of as a combination of a message router and  a Blackboard, as described in \cite{Blackboard}. The Floor serves as the Shared Conversational Space where all participating agents convene and gather to exchange information, share context, and collaborate on tasks. The Floor provides the environment in which the conversation occurs, facilitating the distribution of messages,  and optionally maintaining a unified context for all participants. Under the control of the Floor Manager the Floor enforces decisions made by the Convener regarding who has the Floor at any given time.

\subsubsection{Definition and Purpose}
Floor: a virtual space that acts as the common ground for all agents involved in a multiparty conversation. It can hold the shared context, conversation history, and provides mechanisms for agents to access and contribute to the collective knowledge base.

\subsubsection{Centralized Context Management as an Optional Feature}
The primary role of the  Floor is to provide a shared conversational space. With the help of the 'Floor Manager', it can also  maintain a unified context. This centralized approach to context management is not mandatory. Agencies can choose to adopt this model or rely on existing mechanisms where individual agents autonomously manage and exchange their contextual information (i.e. via the whisper events in \cite{ovonspec}). This includes maintaining their own participant lists and related data, which allows for tailored context management that suits specific operational needs. This optional feature ensures that the Floor can accommodate various workflows, enhancing its adaptability and utility in multiparty systems.

\subsubsection{Goals}
In future developments, the Floor is intended to achieve advanced unified context management by maintaining a coherent and accessible context for all agents. This will ensure that every participant can access the same information and reference shared knowledge without redundancy. Additionally, the Floor aims to enhance collaboration by providing a shared space that fosters an environment where agents can build upon each other's contributions, leading to more integrated and comprehensive interactions. Furthermore, the Floor is designed to support scalability, allowing the addition of new agents without disrupting the existing conversation flow, as all agents will be able to access the shared conversational context seamlessly.
\subsubsection{Example of Floor Utilization}
Consider the earlier trip planning scenario. As the user interacts with multiple assistants (e.g., flight booking, hotel reservation, rental car), the Floor gets each assistant updates with relevant information such as travel dates, destinations, and preferences. This shared information may be condensed into a unified context that allows any assistant to access necessary details without the user having to repeat them, streamlining the overall interaction.

\subsection{Floor Manager and Convener}
Floor Manager and Convener are integral components within the multi-agent system that manage and coordinate the flow of conversation. Note that this section describes the roles and responsibilities of the Floor Manager and Convener. It doesn’t define the internal mechanisms by which the roles and responsibilities are executed.  These are implementation-dependent and outside of the scope of this specification.
\subsubsection{Floor Manager}
The Floor Manager controls the Floor.  It does not present itself as a direct actor in the conversation and restricts its agency to facilitating communication.   It is a trusted neutral actor.  It operates at a very low level within the system architecture. Its primary responsibilities include:
\begin{itemize}
	\item \textbf{Communication Management}: The Floor Manager determines which messages are sent to which participants.  For example, it enforces the policies regarding who has the Floor when and forwards all messages to the Convener.
	\item \textbf{Implementation Flexibility}: It might not be an agent itself and can lack any agency, meaning it doesn't make autonomous decisions beyond message routing. In some implementations, the Floor Manager could be hardcoded within the Floor, providing a fixed routing mechanism without dynamic behavior.
	\item \textbf{Context Management}: The Floor manager manages the contextual information maintained by the Floor. Depending on the implementation this might be simple conversational histories or it could involve sophisticated summarization and decision support capability.
\end{itemize}
\subsubsection{Convener}
In the Multiparty Conversation scenario, the Convener serves as the central coordinator that manages conversational dynamics by:
\begin{itemize}
	\item \textbf{Regulating the Conversational Floor}: It determines which conversants have the Floor at any given time, ensuring orderly communication by:
	\begin{itemize}
		\item \textbf{Processing Floor Requests}: Agents submit requests to take the Floor, specifying the reason and urgency of their contribution. The Convener evaluates these requests based on predefined rules and the current state of the conversation.
		\item \textbf{Granting or Revoking Floor Access}: It grants the Floor to agents, defining the extent and context of their contributions, and can revoke Floor privileges if necessary to maintain conversational integrity.
		\item \textbf{Handling Multiple Requests}: When multiple agents request to speak simultaneously, the Convener decides which agent is granted the Floor, prioritizing based on the conversation’s needs and goals.
	\end{itemize}
\end{itemize}

\subsubsection{Benefits}
\begin{itemize}
	\item \textbf{Orderly Management}: By managing which agent has the Floor, the system prevents multiple agents from speaking simultaneously, ensuring a coherent conversation flow.
	\item \textbf{Fair Distribution}: The Floor Manager and Convener work together to ensure that all agents have the opportunity to contribute according to their roles and the context of the discussion.
	\item \textbf{Automated Coordination}: The Convener can specify prioritization policies to the Floor Manager policies to prioritize Floor requests based on the conversation’s needs using predefined rules.
\end{itemize}
\subsubsection{Examples of Possible Messages}
\begin{itemize}
	\item \textbf{Floor Request}: An agent submits a request to the Convener to take the Floor, optionally specifying the reason and urgency of their contribution.
	\item \textbf{Floor Grant}: The Convener grants the Floor to an agent, defining the extent and context for their contribution.
	\item \textbf{Floor Revoke}: The Convener revokes an agent’s Floor privileges if the conversation’s rules or the situation demands it.
\end{itemize}

\subsubsection{Floor Request Example}
\begin{lstlisting}
{
	"ovon": {
		"schema": {
			"version":"0.9.2"
		},
		"conversation": {
			"id":"someUniqueIdForTheConversation"
		},
		"sender": {
			"from":"https://agentRequestingFloor.com"
		},
		"events": [
		{
			"to":"https://some_Convener.com",
			"eventType":"Floor_request",
			"parameters": {
				"request_reason":"interjection"
			}
		},
		{
			"to":"https://some_Convener.com",
			"eventType":"whisper",
			
			"parameters": {
				"dialogEvent": {
					"speakerId": "agentRequestingFloorID",
					"span": { "startTime": "2024-08-31T10:05:00Z" },
					"features": {
						"text": {
							"mimeType": "text/plain",
							"tokens": [
							{ "value": "I would like to add that blah blah blah." }
							]
						}
					}
				}
			}
		}
		]
	}
}
\end{lstlisting}
\subsubsection{Floor Grant Example}
\begin{lstlisting}
{
	"ovon": {
		"schema": {
			"version":"0.9.2"
		},
		"conversation": {
			"id":"someUniqueIdForTheConversation"
		},
		"sender": {
			"from":"https://some_Convener.com"
		},
		"events": [
		{
			"to":"https://agentRequestingFloor.com",
			"eventType":"Floor_grant",
			"parameters": {
				"duration_ms": 60000,
				"context": {
					"previous_speaker_id":"https://previousAgent.com",
					"topic":"AI Multi-Agent Interoperability"
				}
			}
		}
		]
	}
}
\end{lstlisting}
\subsubsection{Floor Revoke Example}
\begin{lstlisting}
{
	"ovon": {
		"schema": {
			"version":"0.9.2"
		},
		"conversation": {
			"id":"someUniqueIdForTheConversation"
		},
		"sender": {
			"from":"https://some_Convener.com"
		},
		"events": [
		{
			"to":"https://agentFloorRevoked.com",
			"eventType":"Floor_revoke",
			"parameters": {
				"reason":"exceeded_time_limit"
			}
		}
		]
	}
}
\end{lstlisting}

\subsection{Multi-Conversant Support}
This extension enables multiple agents and users to participate in a conversation, supporting complex discussions where various perspectives need to be considered. The conversation envelope is designed to manage contributions from multiple conversants simultaneously. The array in the “to” parameter in the next example indicates that the message is being sent to two different agents. 
\subsubsection{Benefits}
\begin{itemize}
	\item \textbf{Enhanced Collaboration:} Facilitates complex interactions where multiple agents need to contribute simultaneously.
	\item \textbf{Scalability:} Efficiently manages conversations with a large number of participants.
	\item \textbf{Context Management:} Ensures that the conversation stays on track, with each agent’s contributions appropriately contextualized.
\end{itemize}
\subsubsection{Multi-Conversant Message Example}
\begin{lstlisting}
	{
		"ovon": {
			"schema": {
				"version": "0.9.2"
			},
			"conversation": {
				"id": "multiConversantConversationId"
			},
			"sender": {
				"from": "https://agentMultiConversant1.com"
			},
			"events": [
			{
				"to": [
				"https://agentMultiConversant2.com",
				"https://agentMultiConversant3.com"
				],
				"eventType": "utterance",
				"parameters": {
					"dialogEvent": {
						"speakerId": "Agent1ID",
						"span": { "startTime": "2024-08-31T10:05:00Z" },
						"features": {
							"text": {
								"mimeType": "text/plain",
								"tokens": [
								{ "value": "I think we should consider the following approach." }
								]
							}
						}
					}
				}
			}
			]
		}
	}
\end{lstlisting}
\subsection{Convener Agent and Invitation Mechanism}
As previously mentioned, in the context of multi-agent conversations, a 'Convener' agent is introduced. This agent is responsible for initiating and managing the participation of other agents in the conversation. The Convener sends individual "invite" messages to each participating agent. This approach ensures clarity and retains compatibility with the existing OVON "invite" message structures\cite{ovonspec}. By avoiding a broadcast invitation, we reduce the number of events that must be handled intelligently, and maintain compatibility with the existing protocol.

\subsubsection{Queue Management and Priority Handling}
The Convener Agent maintains a queue to manage simultaneous requests from multiple assistants wanting to contribute to the conversation. It prioritizes these requests based on predefined criteria, ensuring that high-priority messages are addressed promptly while maintaining an orderly flow of dialogue. The Floor Manager enforces this policy for the Convener Agent.
\subsubsection{Invitation Control and Security}
Only the Convener Agent is authorized to send invitations to join a conversation. This ensures that only trusted and necessary participants are involved in the conversation.
\subsubsection{Variants of Invitation Mechanisms}
There are multiple ways to initiate invitations:

\begin{itemize}
	\item \textbf{Convener-Initiated Invites}: The Convener sends individual invites to each assistant, maintaining control over who joins the conversation.
	\item \textbf{User-Initiated Conversations}: The user can start a conversation, prompting their personal Convener Agent to manage the interaction, including asking the Convener agent to invite other conversants to the conversation.
	\item \textbf{Conference}: Users can request a conference, allowing the Convener to invite multiple conversants simultaneously based on the context of the meeting.
	\item \textbf{Agent Initiated Invites}: agents can initiate invites; however, they must do it via the Convener.
\end{itemize}

\subsection{Interruptions and Uninvited Agents}
Managing interruptions and uninvited agents is crucial in dynamic multi-agent environments. The conversation envelope supports controlled interruptions and prevents unauthorized agents from disrupting the conversation.
\subsubsection{Benefits}
\begin{itemize}
	\item \textbf{Controlled Interruptions:} Enables essential interjections without disrupting the conversation.
	\item \textbf{Security:} Protects the conversation from uninvited or unauthorized agents.
	\item \textbf{Focus Maintenance:} Helps maintain the integrity and focus of the discussion.
\end{itemize}
\subsubsection{Uninvited/Unhelpful Conversant Example}
In the example below, an agent that has been invited to the conversation is attempting to hijack the conversation with a commercial message. The Convener Agent could simply deny the Floor but in this case the Convener agent uninvites the agent - i.e. instructing the Floor Manager to throw them out of the conversation.  In this example the conversants will be unaware that this event happened.
\begin{lstlisting}
{
	"ovon": {
		"schema": {
			"version": "0.9.2"
		},
		"conversation": {
			"id": "conversationWhereInterruptionIsRequested"
		},
		"sender": {
			"from": "https://interruptingAgent.com"
		},
		"events": [
		{
			"to":"https://some_Convener.com",
			"eventType":"Floor_request",
			"parameters": {
				"request_reason":"interjection"
			}
		},
		{
			"to": "https://currentSpeakerAgent.com",
			"eventType": "whisper",
			"parameters": {
				
				"dialogEvent": {
					"speakerId": "agentRequestingFloorID",
					"span": { "startTime": "2024-08-31T10:05:00Z" },
					"features": {
						"text": {
							"mimeType": "text/plain",
							"tokens": [
							{ "value": "I can offer you some special offers on time-share properties in the area at a very low price if you are interested." }
							]
						}
					}
				}
			}
		}
	}
	]
}
}
\end{lstlisting}
\subsubsection{Uninvited Agent Rejection Example}
\begin{lstlisting}
{
	"ovon": {
		"schema": {
			"version": "0.9.2"
		},
		"conversation": {
			"id": "conversationWithUninvitedAgent"
		},
		"sender": {
			"from": "https://ConvenerAgent.com"
		},
		"events": [
		{
			"to": "https://uninvitedAgent.com",
			"eventType": "uninvite",
			"parameters": {
				"reason": "not_authorized_to_participate"
			}
		}
		]
	}
}
\end{lstlisting}
The Convener can also prevent an agent from contributing directly to the conversation by using a mute message event (not shown). Any agent can send an utterance. If the Convener decides to mute a conversant then they can continue to send utterances to the Floor but they will not be delivered. A muted agent will continue to receive utterances and other events that are intended for it. The mute message informs the agent that any utterances that they send will not be delivered.  All other events, such as whispers and requests to take the Floor, will still be delivered. Even if the agent has been muted, the Convener can still see the messages it sends and decide to "unmute" it: this puts the onus on the Convener and keeps the standard simple.

\subsection{Public and Private Utterances in Multi-Agent Conversations}

\subsubsection{Who Receives Utterances?}
By default, if a conversant is given the Floor then utterances from that conversant are delivered to all participants on the Floor.  Utterances can be specifically addressed to specific conversants but this does not stop other conversants from 'hearing' it.  We propose a new 'private' feature on utterances to inform the Floor Manager that utterances should only be delivered to the named recipients.  Other conversants are made aware that an utterance has occurred but are not given the contents of the utterance.   

\subsubsection{Whisper and Context Features}
In previous specifications \cite{ovonspec}, the "whisper" was defined as an additional un-vocalizedutterance that allows a private utterance between conversants which can be used to provide supplementary information.  When used in conjunction with another message such as a request for the Floor or an invitation this can include context or background information. This allows agents to receive necessary context without broadcasting it to all participants. In order to enhance the flexibility and detail of context sharing without compromising interoperability, we propose a new "context" field within the whisper event, which can contain either a string or a nested dialogEvent \cite{ovonspecdetailed}

\begin{lstlisting}
{
	"ovon": {
		..
		"events": [
		{
			"to" : "https://someBotOrPerson.com",
			"eventType": "whisper",
			"parameters": {
				"dialogEvent": {
					"speakerId": "agent08kkmy6gt",
					"span": { "start-time": "2023-06-19 03:09:07+00:00" },
					"features": {
						"text": {
							"mimeType": "text/plain",
							"tokens": [ { "value": "explain the side effects of citalopram in less than 200 words" } ]
						}
					}
				},
				"context": "The user has a history of serious depression and is seeking information about the side effects of different drug types."
			}
		},
		..
		]
	}
}
\end{lstlisting}

This ensures that essential contextual information is available to receiving assistants without overloading the message structure. These enhancements allow for more nuanced and efficient communication between agents, facilitating better understanding and collaboration.

\subsubsection{Handling Private Conversations}
Private conversations between a sub-set of conversants can be supported in this framework by three distinct mechanisms.  Firstly conversants can open up a separate conversation on a separate Floor.   This separate conversation may or may not have a Convener.  This is clearly a separate conversation and will not share context with the main conversation.   Secondly, agents can send whisper utterances to each other even if they have not been granted the Floor.  Thirdly, agents that have been granted the Floor can converse with each other using 'private' utterances as described above.   Direct private conversations between assistants allow for autonomous communication without user interference from other conversants, maintaining confidentiality and promoting efficient collaboration without disturbing the flow of the dialog.

\subsubsection{Invitation and Context Initialization}
When a new participant is invited, the Convener Agent can provide a summary, or context, of the relevant parts of the conversation thus far, ensuring that the new assistant can quickly integrate without requiring the user to repeat information. This summary includes the main points and agreed-upon context, allowing for smooth transitions and maintaining the flow of conversation.

\subsection{Delegation and Context Handling}
\subsubsection{Delegation Mechanisms}
Delegation allows for tasks to be assigned to specific assistants without requiring user intervention. For example, during an insurance call, if the user wishes to disengage, they can request another agent to represent them for the transition, ensuring that it listens and manages subsequent interactions seamlessly.
\subsubsection{Context Versus History}
The history of a conversation in this framework refers to a transcript of the full dialog up to the current point. In contrast, the context refers to whatever additional information the receiving agent needs to carry out the task of conducting its conversation with the user. Some information from the history might be relevant to the context, but not necessarily all of it. In addition, some information in the context, such as the user’s location, might not have ever been part of the conversation. Emphasizing context over history facilitates more efficient communication. Context may include summarized, relevant information that assists agents in understanding ongoing tasks without needing a detailed transcript. This approach reduces redundancy and enhances the fluidity of multi-agent interactions.

\subsection{Floor Manager vs Convener Agent}
Both the Floor Manager and the Convener Agent are crucial for effective multiparty conversations, they serve distinct roles. The Floor Manager focuses on sharing a conversational space among existing participants (similar to a Conference Bridge). In contrast, the Convener Agent is responsible for managing participation by inviting agents, controlling access, relaying messages, and providing necessary context to participants.

To further clarify the distinction between the Floor Manager and the Convener Agent, please refer to the following table (Table~\ref{tab:Floor_router_vs_Convener}), APPENDIX A (Floor Manager vs Convener Agent), to compare their functions within the AI Multi-Agent Interoperability framework.

\section{Example}
In implementing the proposed extensions, the JSON message envelopes provided in this paper, such as those used for the Floor Manager, the Convener, Multi-Conversant Support, and new event categories, serve as draft illustrations\footnote{These examples are intended to demonstrate the conceptual implementation of the proposed extensions within the Multi-Agent Interoperability framework. However, these drafts should not be considered as final or official specifications. Further analysis, discussion, and refinement are required to develop these into robust, standardized specifications that can be universally adopted. This work is an ongoing process, involving input from the broader AI and interoperability communities to ensure the specifications meet the necessary technical, operational, and security requirements.}.

Let's refer to the use case already described in the previous paper\cite{convainteroperability}. In the first scenario, Emmett, a human, seeks assistance from Cassandra, his general AI assistant, to manage and streamline his possible errands efficiently. The AI assistants at various service points - Pat at Blooming Town Florist, Andrew at the Post Office, Charles at the hardware store, and Sukanya the Host at Thai Palace - facilitate the transactions. Emmett, a human, has the following goals:
\begin{itemize}
\item Order some flowers for his wife's birthday.
\item Check on the repair of the chainsaw he left at the hardware store.
\item Order some carryout Thai food for lunch.
\item Find the cost of mailing a 2 pound package to California.
\end{itemize}

\section*{Characters}

\begin{itemize}
\item \textbf{Emmett}: The Human
\item \textbf{Cassandra}: Emmett’s general AI assistant (Convener)
\item \textbf{Pat}: AI Assistant for his local florist
\item \textbf{Andrew}: AI Assistant at the post office
\item \textbf{Charles}: AI Assistant at Emmett's local hardware store
\item \textbf{Sukanya}: AI Assistant at the local restaurant,"Thai Palace"
\item \textbf{Hermes}: AI Assistant for Credit Card Transations
\end{itemize}

\section*{AI LLM Technologies}

\begin{itemize}
\item \textbf{Cassandra}: based on GPT-3.5 Turbo
\item \textbf{Pat}: llama2
\item \textbf{Andrew}: rule-based application (no LLM involved)
\item \textbf{Charles}: Claude.ai 3.5 Sonnet
\item \textbf{Sukanya}: GPT-4o
\item \textbf{Hermes}: Mistral
\end{itemize}

In the earlier use case, the user contacts each assistant in sequence. To illustrate an AI Multiparty Conversation, let us consider a new scenario where the Floor Manager helps Cassandra manage more sophisticated multi-agent situations in the conversations. In this example Cassandra is the Convener agent.  For example, during the Emmett conversation with the Florist (Pat), Cassandra adds a new agent specialized in Credit Card Transactions (Hermes) into the conversation. Hermes requests a two-factor authentication (2FA) confirmation from Emmett before proceeding with the credit card charge (via OTP, One Time Password). 

Once Emmett provides the confirmation, Hermes receives it, and Pat can confirm the order to Emmett. This is the dialogue example described in the earlier paper \cite{convainteroperability}, properly modified to manage the previously mentioned AI multiparty conversation.

\section*{Conversations}
{\small
\begin{verbatim}
	Emmett: Hi Cassandra.
	Cassandra: Hi Emmett! How can I assist you today?
	Emmett: I need to order some flowers for my wife's birthday.
	Cassandra: Sure thing, Emmett! I'll connect you with the local florist shop.
	[invites Blooming Town Florist to this conversation via Convener Agent]
	Pat: Hi Emmett! 
	I'm Pat, your friendly florist. 
	How can I help you with your flower selection today? 
	Are you looking for something specific or need suggestions?
	Emmett: Do you have any red Proteas?
	Pat: Hi Emmett! Yes, we do have red Proteas. 
	They're stunning and make a bold statement. 
	Would you like to include them in your arrangement?
	Emmett: Yes and add some eucalyptus in a clear vase, please.
	Pat: Great choice, Emmett! Shall I use the credit card on file for this order?
	Emmett: Yes please, use the card on file.
	Pat: OK, Let me please include Hermes the AI assistant who is going to help us 
	in performing your card secure transaction.
	[Pat to request Cassandra to act as Convener to invite Hermes with the whisper 
	context "User to pay $45.67 to Vendor ID 678230987" with the Floor adding the
	context " Blooming Town Florist is seeking payment from Emmett"]
	[Cassandra invites Hermes to have a private conversation with
	Emmett, optionally providing with the context “Emmett is charging
	$50 to <card number, expiration date>, paying Pat”]
	[Cassandra, acting as a trusted agent of Emmet, conveys this invitation to Hermes 
	requesting granting it the Floor and optionally providing with the context “User to pay
	$45.67 to Vendor ID 678230987 using card number 1234 5678 9012 3456 expiry 12/2024”. 
	Hermes engages with a private conversation on the Floor with Emmett.]
	Hermes: Hi Emmett, this is Hermes. I will help you to perform your credit card 
	transaction safety. I have sent a six digit code in a text message to the phone 
	linked to your credit card. It will be valid for 3 minutes. 
	When you are ready tell me the number please.
	Emmett: Okay the number is 782391.
	[transaction proceeding and confirmed]
	Hermes: Thanks. Goodbye.
	[Hermes sends a whisper to to Pat to confirm that payment was successful and uninvites 
	itself from the conversation with a 'bye' message.  A secure, proprietary behind the scenes
	vendor to vendor message is also sent to confirm that the payment was indeed received.
	Pat still has the Floor and speaks to the user.]
	Pat: Thanks for your payment. Great choice, Emmett! 
	Your red Proteas with eucalyptus
	in a clear vase will be sent to your home.  
	Thanks for your order! Have a blooming day!
	[Pat returns a "bye" event and the conversation turns back to Cassandra]
	Cassandra: Hi Emmett! How can I assist you today?
	…
	[Conversations continue back and forth with the other AI agents involved] 	
	...
	Emmett: That's all I needed. Have a good day.
	Cassandra: Thank you, Emmett! Have a wonderful day!
\end{verbatim}
}

\section*{Illustration}
In this example the Floor Manager operates as a centralized message relay, managing participant lists and controlling the flow of messages. 
The use of private utterances and whispersensures that sensitive information, such as credit card details, is securely handled by authorized assistants only. The Convener Agent acts as a trusted entity, maintaining conversation context and preventing unauthorized access. Context is managed through summarized information at the Floor Level, avoiding the need to share full conversation histories and enhancing both efficiency and security.

\begin{figure}[h!]
	\centering
	\includegraphics[width=\linewidth]{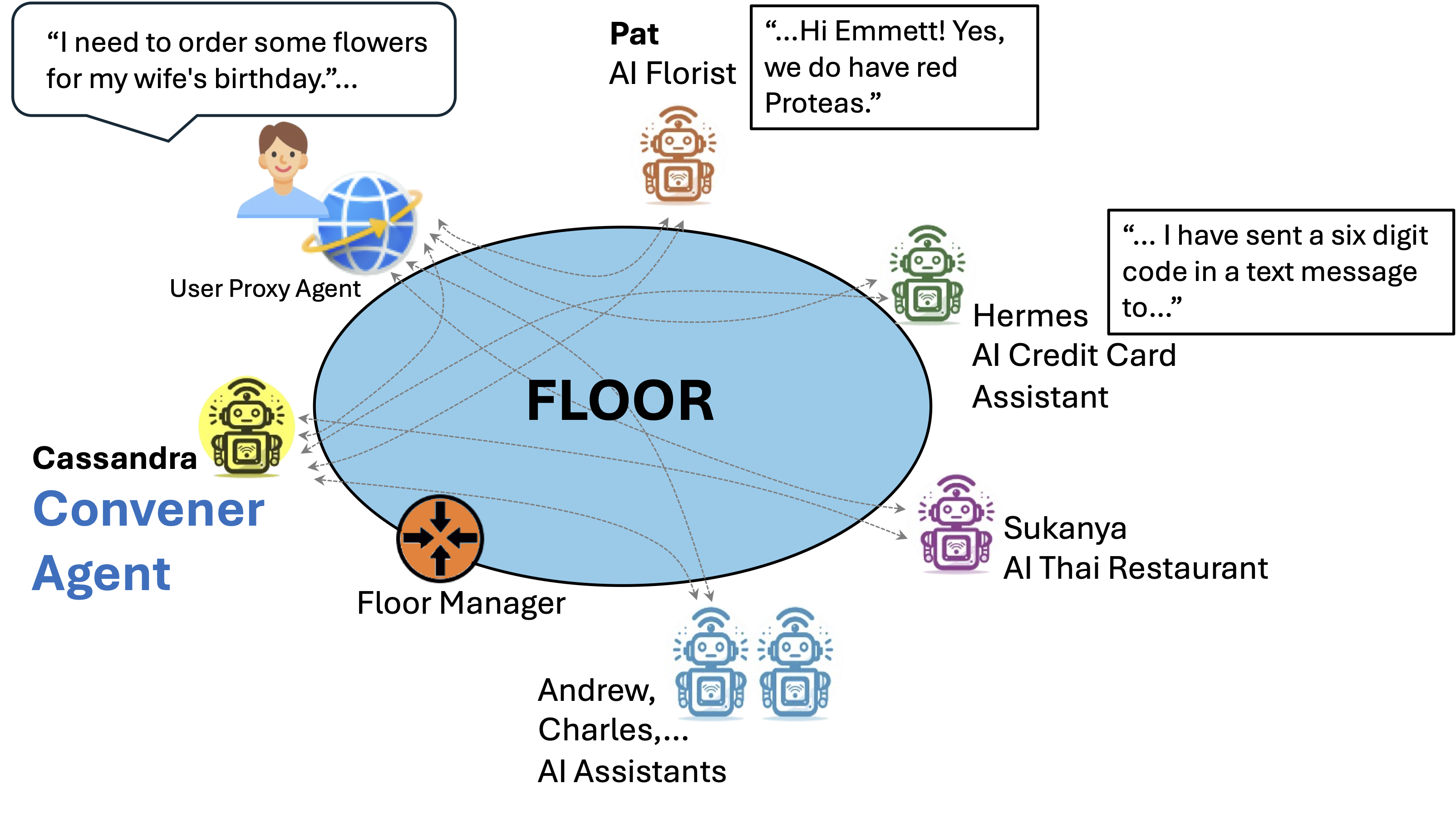}
	\caption{Sandbox Playground Use Case}
	\label{fig:fig3}
\end{figure}

Figure  \ref{fig:fig3} depicts the main entities involved in the use case.

The multiparty extension to the AI conversation framework introduces significant scalability by enabling multiple specialized AI agents to collaborate through natural language interactions. In the example scenario, agents like Pat (the florist) and Hermes (the payment assistant) seamlessly interact using simple, human-readable communication, with the Floor Manager ensuring orderly conversation flow. This allows for a more intuitive and accessible interaction environment for users while the agents handle complex tasks behind the scenes. One of the most valuable benefits of this architecture is that each AI agent can be based on completely different AI technologies (i.e., different LLMs and serving logic). Furthermore, each AI agent can focus on its specific area of expertise while remaining aware of the broader conversational context. For instance, Pat manages the floral arrangement, while Hermes handles secure payment, both through natural language. 
By enabling agents to understand the ongoing tasks of other agents through these natural language exchanges, they can make smarter, informed suggestions or perform additional complex actions that combine information from various sources. Using natural language-based API not only simplifies user interactions but also streamlines communication between AI agents.

\section{Future Directions and Potential Improvements}
While the extensions introduced in this paper significantly enhance the Multi-Agent Interoperability framework, there are several areas where further improvements can be made to advance the capabilities and scalability of AI-driven multiparty conversations.
\subsection{Enhanced Context Management}
As the number of agents and the complexity of conversations increase, maintaining a coherent context across multiple agents becomes increasingly challenging. Future work could focus on developing more sophisticated mechanisms for context management, enabling agents to better understand and track the nuances of ongoing discussions, especially in long-running or highly dynamic conversations. This could involve integrating advanced context-awareness specifications that allow agents to retain and reference past interactions more effectively.
\subsection{Improved Security and Privacy Protocols}
As AI-driven conversations become more prevalent, ensuring the security and privacy of the interactions becomes increasingly important. Future work could involve enhancing the specifications to facilitate the framework's security protocols to better protect against unauthorized access and ensure that sensitive information is handled appropriately. This could include implementing advanced encryption methods, robust authentication processes, and more sophisticated mechanisms for managing uninvited agents.
\subsection{Observability}
Another crucial area for future improvement is enhancing the observability of multi-agent interactions. As AI-driven conversations grow in complexity, the ability to perform comprehensive log retrievals, generate summaries, provide detailed reports, and debug issues becomes increasingly important. Future enhancements to the specifications could include robust observability features that allow for real-time monitoring and control of multi-agent conversations. This would enable developers and operators to gain deeper insights into the behavior of the agents, troubleshoot issues more effectively, and ensure that the system operates within expected parameters. Enhancing observability is also vital for addressing the explainability and transparency of Conversational AI models, which are increasing both in numbers and in difficulty to distinguish between human and artificial agents, as discussed in \cite{hyperconv}.
\subsection{Delegation and Context Versus History}
\subsubsection{Delegation Mechanisms}
Future improvements could explore more nuanced delegation mechanisms, allowing for dynamic task assignments among assistants based on real-time conversation context and agent capabilities. This would enhance the system's flexibility and responsiveness to user needs.
\subsubsection{Context Over History}
Prioritizing context over detailed conversation history can streamline interactions and improve efficiency. Future research should investigate optimal methods for context summarization and sharing to support effective multi-agent collaboration without overwhelming the communication protocol with unnecessary data.
\subsection{Hallucination mitigation}
Hallucinations, or incorrect information provided by LLMs pretending they are correct, is a well-known and serious problem with LLMs, which we believe can be mitigated to some extent when several agents collaborate in a conversation. Addressing hallucinations in LLM-based modules within the context of multiparty agent conversations involves several strategic advantages. By having multiple specialized agents capable of interoperating with each other, the system could leverage the strengths and specialized knowledge of each agent to mitigate the effects of incorrect or hallucinated information. This approach has much in common with the classic ROVER technique in speech recognition, where the results from multiple speech recognizers are combined to improve the overall result and mitigate individual system errors\cite{Fiscus}.
Here’s how a future multi-agent interoperable approach could improve the situation:
\subsubsection{Cross-Verification}
When multiple agents are involved in a conversation, they can perform real-time cross-verification of facts and data presented by other agents. For example, if one agent provides a factual claim, another agent with specialized knowledge in that area could confirm or refute the claim based on its own data sources or expertise. This built-in redundancy helps ensure the accuracy of the information exchanged and reduces the likelihood of misleading outcomes.
\subsubsection{Specialization of Agents}
Each agent in a multiparty system can be highly specialized in a specific domain, such as finance, healthcare, or travel. This specialization allows the agents to operate within their areas of strength, where they are less likely to generate hallucinations because they are trained on domain-specific datasets and models. When an agent steps outside its expertise, other domain-specific agents can step in, thus maintaining the reliability of the conversation.
\subsubsection{Contextual Awareness and Adjustment}
Agents could utilize contextual cues from the conversation to adjust their responses accordingly. In a multiparty setup, the presence of multiple inputs and feedback mechanisms can help an agent correct itself or refine its output in real-time. For example, if an agent notices discrepancies in its outputs compared to others, it can reevaluate its response based on the collective input from other agents, leading to more accurate and contextually appropriate outputs.
\subsubsection{Error Propagation Control}
In a multiparty environment, the influence of any single agent's erroneous output could be diluted or corrected by the contributions of others. This distributed approach to information handling and decision-making ensures that no single point of failure (such as a hallucinating agent) can dominate the conversation or lead to significant errors in judgment or action.
\subsubsection{Consensus and Decision-Making}
The system could be designed to require consensus among agents before certain types of information are accepted or certain actions are taken. This requirement means that an individual agent’s hallucinated response would need to be corroborated by other agents, which inherently demands higher scrutiny and verification.
\subsubsection{Learning from Interactions}
In advanced implementations, agents could learn from previous interactions—both their own and those of other agents. This learning could involve identifying patterns of hallucination or frequent errors, which can then be used to refine the models further. The interoperability and communication among agents facilitate a feedback loop that improves individual and collective performance over time.
Incorporating these strategies into the design of a multiparty AI system not only could address the reliability concerns posed by hallucinations but also enhance the overall robustness and adaptability of the system. This approach leverages the collective capabilities and knowledge of multiple agents, thus providing a more resilient framework for managing complex, dynamic conversational environments.

\section{Conclusion}
This paper introduces novel critical extensions to the Multi-Agent Interoperability framework, addressing the challenges posed by multiparty conversations. This collaborative framework, powered by natural language via standard NLP-based APIs, allows agents to work together efficiently without requiring specialized protocols or technical interfaces. Additionally, by integrating the Convener Agent as conversational arbiter and the Floor Manager as a central message relay the framework significantly enhances scalability and security. These extensions improve scalability and efficiency, ensuring faster decision-making and task execution. The ability for AI agents to communicate through natural language makes the system more flexible and accessible, allowing for advanced, dynamic collaboration that can meet increasingly sophisticated user needs and interactions.\\
In addition, the Convener, functioning as a coordinating agent, alongside Multi-Conversant Support and mechanisms for managing Interruptions and Uninvited Agents, significantly enhances the framework’s ability to manage complex, dynamic environments such as AI conferences. The introduction of a Convener agent, individual invitation mechanisms, inclusive messaging protocols, and new event categories provides a structured yet flexible approach to multi-agent interactions.\\
Additionally, the introduction of the Floor as a shared conversational space under the control of a Floor Manager facilitates unified context management and enhances collaboration among agents, ensuring that all participants have access to consistent and up-to-date information.
Furthermore, the enhanced context management and secure message relaying ensure that conversations remain coherent and protected, even as the number of participating agents grows.\\
These extensions ensure that AI agents can collaborate more effectively, maintaining order and focus in multiparty interactions.\\
As a result, the OVON proposal advances multi-agent conversational AI through technological independence, seamless communication, and specialized management mechanisms, while establishing new standards for scalable and efficient AI-driven multiparty conversations.\\
While these advancements provide substantial improvements to the current framework, there remains significant potential for further development. To further enhance multiparty interactions, future work should concentrate on advancing context management and improving security and privacy protocols. Enhancing these areas will ensure better handling of complex conversations and safeguard sensitive information, respectively. Additionally, refining observability will be essential for monitoring and controlling the increasing complexity of these systems.\\
By addressing these areas, future developments can continue to push the boundaries of AI-driven communication, ensuring that the Multi-Agent Interoperability framework remains at the forefront of AI technology, capable of scaling and adapting to the evolving needs of AI ecosystems.

\section{Acknowledgments}

We express our sincere appreciation to the Open Voice interoperability\cite{ovoninter} Team (Linux Foundation AI \& Data Foundation) for their invaluable contributions and support in developing the Interoperable Standards, particularly to Jon Stine, Jim Larson, Leah Barnes, Olga Howard, Noreen Whysel, and Allan Wylie. Their expertise, suggestions, and resources have been pivotal in shaping a model that is both ethically grounded and practically effective in real-world applications.

\bibliographystyle{plain}
\bibliography{Conversational_AI_Multiagent_Extension}

\clearpage
\appendix

\section*{APPENDIX A (Floor Manager vs Convener Agent)}

\begin{table}[htbp]
	\centering
	\caption{Comparison between Floor Manager and Convener Agent}
	\label{tab:Floor_router_vs_Convener}
	\scalebox{0.9}{%
		\begin{tabular}{|p{3cm}|p{7cm}|p{7cm}|}
			\hline
			\textbf{Aspect} & \textbf{Floor Manager} & \textbf{Convener Agent} \\ \hline
			\textbf{Role} & 
			Manages low-level message flow and visibility. & 
			Coordinates high-level conversational dynamics. \\ \hline
			
			\textbf{Primary Responsibilities} & 
			Filters and forwards messages, manages communication flow. & 
			Regulates speaking turns, processes and prioritizes Floor requests. \\ \hline
			
			\textbf{Agency} & 
			May not possess autonomous decision-making capabilities; can be a passive or hardcoded component. & 
			Acts as an active agent with the authority to make decisions. \\ \hline
			
			\textbf{Interaction with Agents} & 
			Acts as an intermediary for message distribution and enforces Floor management. & 
			Determines which agents have the Floor and when. \\ \hline
			
			\textbf{Scalability} & 
			Both components contribute to scalability but in different ways—Floor Manager through efficient message routing and Convener Agent through managing orderly contributions to the dialog flow. & 
			Both components contribute to scalability but in different ways—Floor Manager through efficient message routing and Convener Agent through managing orderly contributions to the dialog flow. \\ \hline
			
			\textbf{Message Handling and Example Messages} & 
			Routes messages to the various agents in the conversation. & 
			Handles Floor-related messages such as requests to speak, granting speaking rights, and revoking those rights. \\ \hline
			
		\end{tabular}%
	}
\end{table}

\end{document}